\documentclass[runningheads]{llncs}

\usepackage{multirow}
\usepackage{subfigure}
\usepackage{booktabs}
\usepackage{rotating}

\usepackage{pifont}
\newcommand{\cmark}{\ding{51}}
\newcommand{\xmark}{\ding{55}}

\newcommand{\repeatthanks}{\textsuperscript{\thefootnote}}
\usepackage{graphicx}
\usepackage[
	colorlinks=true,
	urlcolor=blue,
	linkcolor=blue,
	citecolor=blue
]{hyperref}
\usepackage{booktabs}
\usepackage{multirow}
\begin{document}

\title{Making Neural Networks FAIR}

\author{Anna Nguyen\thanks{These authors contributed equally to the work.}\inst{1} \and 
Tobias Weller\repeatthanks\inst{1} \and Michael Färber\inst{1} \and York Sure-Vetter\inst{2}} 

\authorrunning{A. Nguyen et al.}

\institute{Karlsruhe Institute of Technology (KIT), Karlsruhe, Germany \and National Research Data Infrastructure (NFDI), Karlsruhe, Germany}

\maketitle

\begin{abstract}
Research on neural networks has gained significant momentum over the past few years. 
Because training is a resource-intensive process and training data cannot always be made available to everyone, there has been a trend to reuse pre-trained neural networks. As such, neural networks themselves have become research data.
In this paper, we first present the neural network ontology \textit{FAIRnets Ontology}, an ontology to make existing neural network models \textit{findable}, \textit{accessible}, \textit{interoperable}, and \textit{reusable} according to the FAIR principles. Our ontology allows us to model neural networks on a meta-level in a structured way, including the representation of all network layers and their characteristics.
Secondly, we have modeled over 18,400 neural networks from GitHub based on this ontology, which we provide to the public as a knowledge graph called \textit{FAIRnets}, ready to be used for recommending suitable neural networks to data scientists.
\end{abstract}

\section{Introduction}

Researchers of various sciences and data analysts reuse but also re-train neural network models according to their needs.\footnote{\url{https://trends.google.com/trends/explore?date=all&q=transfer\%20learning}} Providing pre-trained neural network models online has the following advantages. First, as a provider, you can benefit from users improving your neural network and circulating your research. Second, as a user of an already trained neural network, you can overcome the cold start problem as well as save on training time and costs. 
Furthermore, providing trained neural network models gets increasingly important in the light of the research community efforts to make research results more transparent and explainable (see FAIR principles~\cite{wilkinson_fair_2016}). As a result, more and more trained models are provided online at source code repositories such as GitHub. The models provided serve not only to reproduce the results but also to interpret them (e.g., by comparing similar neural network models).
Lastly, providing and using pre-trained models gets increasingly important via transfer learning in other domains.

To ensure the high-quality reuse of data sets and infrastructure, the \textit{FAIR Guiding Principles for scientific data management and stewardship}~\cite{wilkinson_fair_2016} have been proposed. These guidelines are designed to make digital assets \textbf{F}indable, \textbf{A}ccessible, \textbf{I}nteroperable, and \textbf{R}e-usable. They have been widely accepted by several scientific communities nowadays (e.g.,~\cite{WISE2019933}).
Making digital assets FAIR is essential to deal with a data-driven world and thus keeping pace with an increasing volume, complexity, and creation speed of data.
So far, the FAIR principles have been mainly applied when providing data sets and code \cite{WISE2019933,DevarakondaPGK19}, but not machine learning models, such as neural network models. In this paper, we bring the FAIR principles to neural networks by (1)~proposing a novel schema (i.e., ontology) which enables semantic annotations to enhance the information basis (e.g., for search and reasoning purposes) and (2)~representing a wide range of existing neural network models with this schema in a FAIR way.
As we outline in Sec.~\ref{sec:DatasetCreation}, extracting metadata from neural networks automatically is a nontrivial task due to heterogeneous code styles, dynamic coding, and varying versioning.
The key idea is that the information contained in these networks should be provided according to the FAIR principles.
This comprises several steps which not only consist of having identifiers but providing (meta)data in a machine-readable way in order to enable researchers and practitioners (e.g., data scientists) easy access to the data. We facilitate this by using semantic web technologies such as OWL and RDF/RDFS. 

Overall, we provide the following contributions:
\begin{enumerate}
 \item We provide an \textit{ontology}, called \textsc{FAIRnets Ontology},
 for representing neural networks. 
 It is made available using a persistent URI by w3id and registered at the platform Linked Open Vocabularies (LOV).\\
 \textbf{Ontology URI:} \url{https://w3id.org/nno/ontology}\\
 \textbf{LOV:} \url{https://lov.linkeddata.es/dataset/lov/vocabs/nno}
 \item We provide a \textit{knowledge graph}, called \textsc{FAIRnets}, representing over 18,400 publicly available neural networks, following the FAIR principles. \textsc{FAIRnets} is available using a persistent URI by w3id and is uploaded to Zenodo.\\
 \textbf{Knowledge Graph URI:} \url{https://w3id.org/nno/data}\\
 \textbf{Zenodo:} \url{https://doi.org/10.5281/zenodo.3885249}
\end{enumerate}
Our contribution is beneficial in several application areas (see Sec.~\ref{sec:impact}). For instance, we already provide an online search system called \textsc{FAIRnets Search} \cite{NguyenW19} by which users can explore and analyze neural network models. 

The paper is structured as follows. 
Sec.~\ref{sec:ontology} describes the structure of \textsc{FAIRnets Ontology} and Sec.~\ref{sec:search_engine} describes the knowledge graph \textsc{FAIRnets}.
Sec.~\ref{sec:fair} explains the reason why the neural networks in \textsc{FAIRnets} follow the FAIR principles.
Sec.~\ref{sec:impact} describes the impact of \textsc{FAIRnets}. 
Sec.~\ref{sec:related_work} gives an overview of related work. 
Lastly, the contributions are summarized.
\section{FAIRnets Ontology}\label{sec:ontology}

\subsection{Creation Process}
\vspace{-0.5cm}
\begin{center}
  \begin{sidewaysfigure*}
   \includegraphics[width=1.1\textwidth]{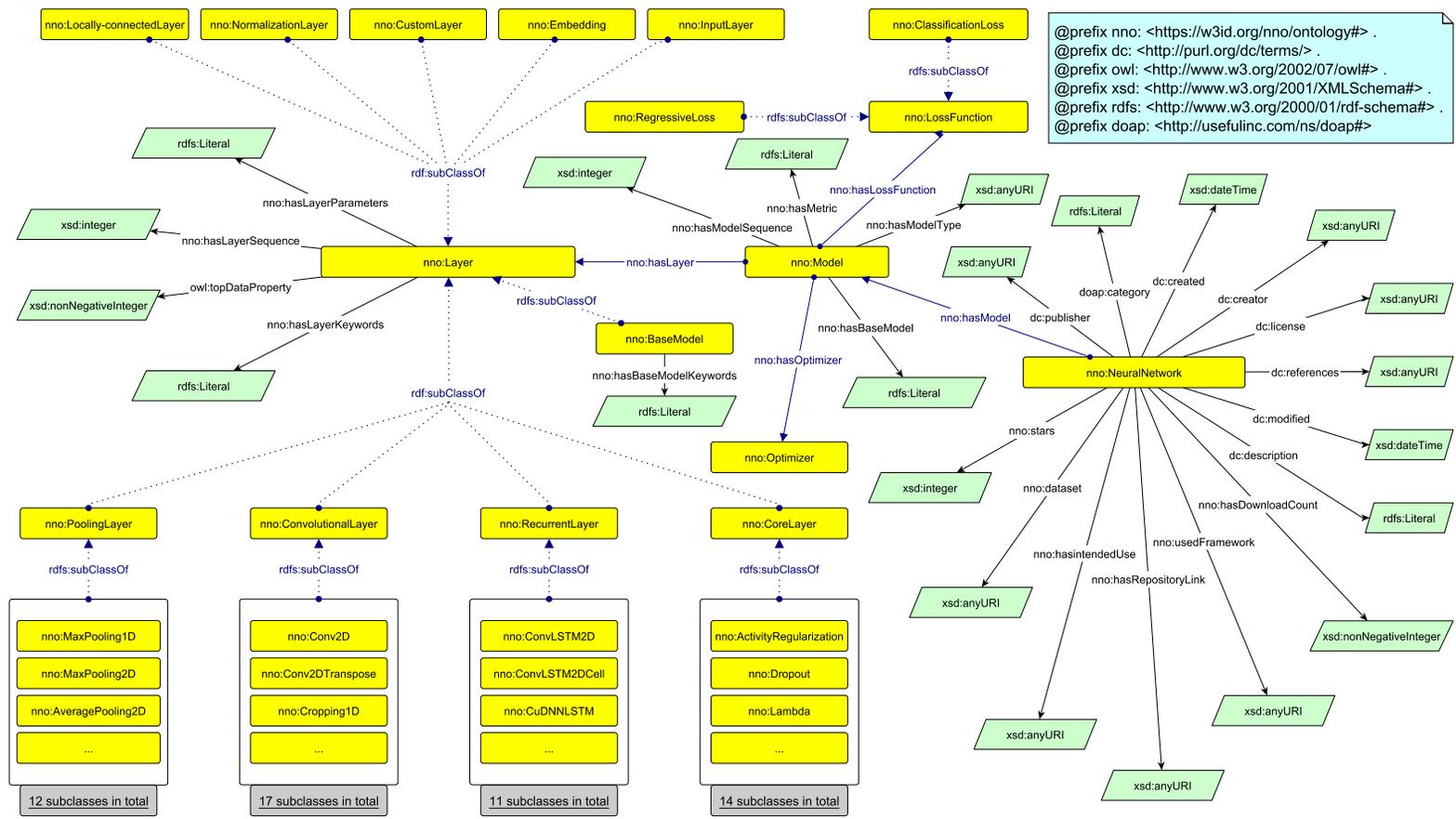}
\caption{Visualization of \textsc{FAIRnets Ontology}.}
\label{fig:UMLClass}
\end{sidewaysfigure*}  
\end{center}
The \textsc{FAIRnets Ontology} is dedicated to model metadata for neural network models on a schema level. 
We developed the ontology by using Prot\'{e}g\'{e}~\cite{musen_protege_2015}. 
To the best of our knowledge, there is no existing vocabulary for the specific description of neural networks. That is why several senior researchers use best practices \cite{GangemiP09} to construct the ontology. We identify researchers, especially beginners, as potential users. The use cases we envision can be found in Sec.~\ref{sec:impact}. 

In addition to the consideration of the Predictive Model Markup Language (PMML) in the development of the ontology (especially in describing the architecture), findings from further work were also considered. 
In particular, model cards~\cite{mitchell_model_2019} were taken into account to validate relevant concepts. 
Model cards encourage transparent model reports concerning machine learning models and are used for outlining the intended use of a model.
These cards define minimal information needed to sufficiently describe a machine learning model (in our case, a neural network) that is relevant to the intended application domains. As suggested from model cards, we included model details such as person developing model, model date, model type, and licenses.

\paragraph{Characteristics.} The structure of the \textsc{FAIRnets Ontology} can be seen in Fig.~\ref{fig:UMLClass}. Overall, the ontology consists of a total of 516 axioms and uses a total of 77 classes where 70 are sub-classes. It also consists of four object properties, 23 data properties, and 29 individuals.

The ontology enables representing three different aspects of information. 
(1) Neural network-related general metadata and
(2) neural network-dependent features can be modeled, such as the type of layer, loss function, and optimizer.
(3) Layer-specific metadata is used to enhance the information basis of the specific layers, e.g., its keywords and parameters.
In the following, we will describe these three components of the \textsc{FAIRnets Ontology} correspondingly.\footnote{We will use \texttt{nno} as the prefix for the namespace \texttt{https://w3id.org/nno/ontology\#}} 

\noindent\textbf{General information}
describe general components of the neural network, as well as the intended use.
For instance, the owner/developer of the (trained) neural network is modeled by using the property \texttt{dc:creator}. This attribute makes it possible to search for repositories by the author in the domain of neural networks.
Following the Linked Data Principles, the author is represented via a URI. 
In this way, the authors are uniquely identified. Therefore, it is possible to link it to the Microsoft Academic Knowledge Graph \cite{Farber19} which models scholarly data such as scientific publications in which some of the represented neural network models are proposed and evaluated.
Moreover, a name (\texttt{rdfs:label}) and a description (\texttt{dc:description}) of the trained neural network are stored. The data property \texttt{nno:dataset} of type URI
allows us to specify the data set that was used to train the neural network. 
This information already gives a more detailed insight into the neural network as well as the intended use of it. 

Furthermore, the timestamp of creation date (\texttt{dc:created}) or last modification (\texttt{dc:modified}) allows assessing the currency of the neural network. \texttt{dc:license} indicates the rights to modify and redistribute that network.
Besides, the property \texttt{nno:hasRepositoryLink} allows linking to the repository in which the neural network is located.
Likewise, references to published papers can be included using \texttt{dc:references}.

\noindent\textbf{Model-specific information}
covers model-specific components of the neural network, such as optimization function denoted by \texttt{nno:hasOptimizer}.
The ontology covers modeling various loss functions, such as binary cross-entropy and mean squared error, via the property \texttt{nno:hasLossFunction}.
Loss functions are subdivided into classification and regression loss functions in the ontology to further indicate the intended use of the neural network. 
The information about existing layers of the neural network can be linked via the property \texttt{nno:hasLayer}.
The loss functions and layer types available in Keras, an open-source deep learning framework to model neural networks, served as a basis to model available loss functions and layers.

\noindent\textbf{Layer-specific metadata} outline additional information about the individual layer. The layers of neural networks are subdivided into subclasses such as core, recurrent, and convolutional layer. 
These classes are further subdivided into more specific layer classes. This specification derived from Keras enables to categorize the neural networks. For example, a neural network with a layer from class convolutional layer can be assigned to the type convolutional neural network.
Furthermore, the hyperparameters (e.g., kernel size, stride, and padding) are denoted by \texttt{nno:hasLayerKeywords} and saved as a dictionary. Additional values in the layer are denoted by \texttt{nno:hasLayerParameter}. 

Most of the categories, properties, and instances are annotated with a label (\texttt{rdfs:label}), a description (\texttt{rdfs:comment}), and, if given, a link (\texttt{rdfs:seeAlso}) which make it easy for ontology users to identify the intended use of categories, properties, and instances, therefore supporting the reusability.

\subsection{Provisioning}
The World Wide Web Consortium (W3C) Permanent Identifier Community Group service is used to provide secure and permanent URL forwarding to the ontology. The \textsc{FAIRnets Ontology} in syntax turtle is accessible under \url{https://w3id.org/nno/ontology}.
Moreover, the ontology has been registered at LOV.\footnote{\url{https://lov.linkeddata.es/dataset/lov/vocabs/nno}, last acc. 2020-10-15}
The ontology is licensed under Creative Commons BY 4.0\footnote{\label{CC}\url{https://creativecommons.org/licenses/by/4.0/}, last acc. 2020-10-15} which allows its wide usage. 
Furthermore, the ontology follows the 5-Star Linked Data Principles\footnote{\label{linkeddata}\url{https://5stardata.info/en/}, last acc. 2020-10-15} and can, therefore, be easily reused. A VoID file is provided under \url{https://w3id.org/nno/fairnetsvoid} including provisioning information.


\section{FAIRnets Knowledge Graph}
\label{sec:search_engine}

Apart from the ontology, we provide the \textsc{FAIRnets} knowledge graph, which is based on the \textsc{FAIRnets Ontology}. 
The knowledge graph allows us to store knowledge (in our case, detailed metadata for neural network models) intuitively as a graph. Existing and widely used W3C standards and recommendations, such as RDF and SPARQL, can be used to query the knowledge graph and to integrate relatively easily into existing frameworks and systems. 
For instance, \textsc{FAIRnets} is already integrated into KBox~\cite{kbox} which is a data management framework, allowing users to share resources among different applications.

\subsection{Creation Process}
\label{sec:DatasetCreation}

The previous online available neural network repositories such as Keras,\footnote{\label{keras}\url{https://keras.io}, last acc. 2020-10-15} Caffe Model Zoo,\footnote{\label{modelzoo}\url{https://github.com/BVLC/caffe/wiki/Model-Zoo}, last acc. 2020-10-15} and
Wolfram Alpha\footnote{\label{wolfram}\url{https://resources.wolframcloud.com/NeuralNetRepository}, last acc. 2020-10-15} are rather small (under one hundred neural networks) and not sufficient to present trends in the development and usage of neural networks. 
General-purpose online code-sharing services, such as GitHub\footnote{\url{https://www.github.com}, last acc. 2020-10-15} and Bitbucket,\footnote{\url{https://www.bitbucket.org}, last acc. 2020-10-15} in contrast, contain many repositories of different nature. We, thus, decided to use GitHub since it is the largest host of repositories. Details about the nontrivial extraction process are given in the following. 

\paragraph{Data Source.}
We extract and represent metadata of publicly available, trained neural network models in RDF* (i.e. RDF and RDFS) based on the \textsc{FAIRnets Ontology}.
Information from SemanGit~\cite{semangit} and GHTorrent~\cite{ghtorrent} can be used to identify GitHub repositories. SemanGit and GHTorrent provide a collection of data extracted from GitHub. In total there are more than 119 million repositories available in the GHTorrent data collection. 
However, SemanGit and GHTorrent have a different focus and do not provide all the information which we wanted to provide in the \textsc{FAIRnets} knowledge graph. For instance, information about the architectures of neural networks within the repositories, the creation date of the repositories, as well as the watcher count is not included. We, therefore, directly accessed the GitHub Repository API and queried available neural network repositories.
We used the search term `neural network' and filtered for repositories that use Python as a programming language.
We accessed these repositories\footnote{Exemplary GitHub API Request: \url{https://api.github.com/repos/dmnelson/sentiment-analysis-imdb}, last acc. 2020-10-15.\label{note:example_request}} and extracted the neural network metadata. 

\paragraph{Extraction Process.}
The difficulty lies in the extraction of the architecture information from the code. We narrowed our extraction down on neural networks implemented in Python. Still, it is difficult to identify the Python file which models a neural network. Therefore, we started with h5 files which are an open-source technology for storing trained machine learning models. Neural networks that have been trained with Keras, for example, can be stored in this format. The h5 file contains information about the neural network, e.g., the sequence of layers, used activation functions, optimization function, and loss function. Accessing the information in the h5 file makes it easier to identify and extract the architecture of the neural network.
However, not every repository contains trained neural networks in h5 files. The reason is that trained neural networks often take up a lot of storage space. Thus, our contribution is the information extraction from the code directly which will be described below. 

{
\begin{table}[tbp]
\centering
\caption{Mapping of GitHub REST API values to the general components in \textsc{FAIRnets}.}
\begin{tabular}{p{3.8cm}p{5cm}}\toprule
\textbf{GitHub API} & \textbf{FAIRnets Ontology} \\ \midrule
created\_at & \texttt{dc:created} \\
description, readme & \texttt{dc:description} \\
html\_url & \texttt{nno:hasRepositoryLink} \\
license & \texttt{dc:license} \\
owner[`html\_url'] & \texttt{dc:creator} \\
updated\_at & \texttt{dc:modified} \\
watchers\_count & \texttt{nno:stars} \\
name & \texttt{rdfs:label} \\
topics[`names'] & \texttt{doap:category} \\
\bottomrule
\end{tabular}
\label{tab:mapping}
\end{table}
}

\noindent\textbf{General Information:}
The mapping of the values from the Github API with the corresponding general component properties in \textsc{FAIRnets} can be seen in Tab.~\ref{tab:mapping}.
We use the \textit{full\_name} of the GitHub REST API as a unique identifier (e.g., `dmnelson/sentiment-analysis-imdb' in note \ref{note:example_request}).
The \textit{full\_name} consists of the GitHub username combined with the name of the repository. 
The owner of the repository is also the owner of the neural network. 
Moreover, we store the link (\texttt{nno:hasRepositoryLink}), the time of creation (\texttt{dc:created}), and the last modification (\texttt{dc:modified}) of the repository. As a description of the neural network (\texttt{dc:description}), we extracted automatically the description and readme file of the GitHub repository. This gives a summary of the possible use of the neural network.
Furthermore, license information about the neural network is extracted and modeled in the knowledge graph, if available. 
This information is crucial regarding the reusability of neural networks.
Given this information, it is possible to filter neural networks by license -- which is often an important constraint in industrial settings.   
To enrich the knowledge graph \textsc{FAIRnets} with information according to the usage of a neural network, we extract the topics\footnote{\url{https://developer.github.com/v3/repos/\#list-all-topics-for-a-repository}, last acc. 2020-10-15.} of each repository from the GitHub repositories and store them as \texttt{doap:category}.

Additionally, we extract arXiv HTTP links within the readme file and map them to \texttt{dc:references}. If BibTex file codes can be found in the readme file, we extract the URL information from the BibTex entry and link it by using the property \texttt{dc:references}. The property \texttt{dc:references} is only intended for scientific contributions. By linking it with URLs from BibTex entries and arXiv links, we ensure this condition. Other links in the readme file are linked to the neural network using \texttt{rdfs:seeAlso}. 

\noindent\textbf{Model \& Technical Information:}
The main feature of \textsc{FAIRnets} is the modeling of neural networks. We can model the structure and technical components of neural networks by employing the \textsc{FAIRnets Ontology}.
To extract the neural network information from the repositories we consider all Python files in the repositories. Each repository can contain several models of a neural network. 
In general, it is difficult to extract the architecture information automatically without executing the source code. By executing the code, you can save the neural network model, for example in h5, and retrieve the information easier. We seek a more elegant way by saving execution costs and use language processing to extract the information.
Due to that, we focus on Python files with static variables.
Despite this restriction, there are still challenges because of various programming styles such as inconsistent naming of variables, complex loop constructions, different structures of code, and other logic statements. Another challenge is changing parameter naming due to different framework versions which are usually not stated.
To solve these tasks, a general method is generated using Python Abstract Syntax Trees (AST) module.\footnote{\url{https://docs.python.org/3/library/ast.html}, last acc. 2020-10-15} The AST module helps Python applications to process trees of the Python abstract syntax grammar. We focused on Keras applications of neural networks to extract the architecture because it is the most used deep learning framework among the top-winners on Kaggle.\textsuperscript{\ref{keras}}
The information on the architecture of the neural network is then modeled by using the schema and properties provided by the \textsc{FAIRnets Ontology}. 
Also, the individual layers and their hyperparameters are stored in our knowledge graph. Likewise, the used optimization function and loss function are stored, among other things, allowing us to infer whether the neural network is used for classification or regression. 
Our code can be found on GitHub.\footnote{\url{https://github.com/annugyen/FAIRnets}} 

\paragraph{Evaluation.}
To evaluate the accuracy of our information extraction, we manually went through 50 examples where we judged the extraction of the GitHub Repository API in Tab.~\ref{tab:mapping}. The evaluation was in all cases correct. In the case of the neural network architecture, we used the h5 files, if available, in the repositories. 
We were able to evaluate over 1,343 h5 files with architecture information (i.e., layer information) which overlap with the architecture extracted from the code with 54\% accuracy. Due to later modifications in the code it is possible that the overlap with the h5 file does not apply anymore (e.g., if a layer is commented out).

\subsection{Provisioning} 
Just like the \textsc{FAIRnets Ontology}, the knowledge graph \textsc{FAIRnets} is also based on the 5-Star Linked Data Principles. 
The knowledge graph is accessible under a persistent URI from w3id and additionally provided on Zenodo. 
In combining FAIR principles and Linked Data Principles using URIs to identify things, providing information using RDF*, and linking to other URIs, it is possible to easily reference and use \textsc{FAIRnets} (see Sec.~\ref{sec:impact}). Machine-readable metadata allows us to describe and search for neural networks. 
The knowledge graph \textsc{FAIRnets}, like the \textsc{Ontology}, is published under the Creative Commons BY 4.0 license.\textsuperscript{\ref{CC}}
A VoID file describing the knowledge graph in a machine-readable format is provided under \url{https://w3id.org/nno/fairnetsvoid}.

\subsection{Statistical Analysis of the FAIRnets Knowledge Graph}
Tab.~\ref{tab:numbers} shows some key figures about the created knowledge graph. 
It consists of 18,463 neural networks, retrieved from 9,516 repositories, and provided by 8,637 unique users. 
The creation time of the neural networks in our knowledge graph ranges from January 2015 to June 2019. All these networks have a link to the respective repository and owner. 
Based on the used layers, we can infer the type of neural network. If a network uses a convolutional layer, it is inferred that the network is a convolutional neural network (CNN). Likewise, if a network contains a recurrent layer, it is inferred that the network is a recurrent neural network (RNN). For simplicity, if none of those two layer types are used, the default claim for the network is a feed-forward neural network (FFNN). 
Of the total 18,463 neural networks, FFNN is most represented in the knowledge graph comprising half of the neural networks. 
CNNs follows with 36\% and RNN with 16\% of the total number of neural networks. 

{
\begin{table}[tbp]
\centering
\caption{Statistical key figures about \textsc{FAIRnets}.}
\begin{tabular}{p{4cm}p{3cm}}\toprule
\textbf{Key Figure} & \textbf{Value} \\ \midrule
repositories  & 9,516 \\
unique users & 8,637 \\ 
neural networks & 18,463 \\
\hspace{0.2cm} FFNN & 8,924 (48\%)\\
\hspace{0.2cm} CNN & 6,667 (36\%)\\
\hspace{0.2cm} RNN & 2,872 (16\%)\\ 
\bottomrule
\end{tabular}
\label{tab:numbers}
\end{table}
}

\section{FAIR Principles for Neural Networks}
\label{sec:fair}

\begin{table*}[tbp]\centering
\caption{Evaluation of \textsc{FAIRnets} according to the Generation2 FAIRMetrics (Note: \cmark = passed, (\cmark)= should pass, \xmark = not passed).}
\label{tab:fairmetrics}
\begin{small}
\begin{tabular}{@{}lp{2.5cm}p{6.5cm}cp{1cm}@{}}\toprule
Principle & FAIRMetric & Name & Result \\ \midrule
\multirow{5}{*}{Findable} & \textit{Gen2\_FM\_F1A} & Identifier Uniqueness & \cmark \\
& \textit{Gen2\_FM\_F1B} & Identifier persistence & \cmark \\
& \textit{Gen2\_FM\_F2} & Machine-readability of metadata & \cmark \\
& \textit{Gen2\_FM\_F3} & Resource Identifier in Metadata & \cmark \\
& \textit{Gen2\_FM\_F4} & Indexed in a searchable resource & \xmark \\
\midrule
\multirow{4}{*}{Accessible} & \textit{Gen2\_FM\_A1.1} & Access Protocol & \cmark \\
& \textit{Gen2\_FM\_A1.2} & Access authorization & \cmark \\
& \textit{Gen2\_FM\_A2} & Metadata Longevity & (\cmark)  \\
\midrule
\multirow{3}{*}{Interoperable} & \textit{Gen2\_FM\_I1} & Use a Knowledge Representation Language & \cmark \\
& \textit{Gen2\_FM\_I2} & Use FAIR Vocabularies & \cmark \\
& \textit{Gen2\_FM\_I3} & Use Qualified References & \cmark \\
\midrule
\multirow{3}{*}{Reusable} & \textit{Gen2\_FM\_R1.1} & Accessible Usage License & \cmark \\
& \textit{Gen2\_FM\_R1.2} & Detailed Provenance & (\cmark) \\
& \textit{Gen2\_FM\_R1.3} & Meets Community Standards & (\cmark) \\
\bottomrule
\end{tabular}
\end{small}
\end{table*}

With \textsc{FAIRnets}, we treat neural networks as research data. As such, to ensure good scientific practice, it should be provided according to the FAIR principles, that is, the data should be findable, accessible, interoperable, and reusable. 
While the GitHub repositories themselves do not satisfy the FAIR principles (e.g., the metadata is not easily searchable and processable by machines), the modeling of the neural networks in the \textsc{FAIRnets} knowledge graph is made \textit{FAIR} as we show in the following.
Specifically, in this section, we identify the factors that make the neural network representations in \textsc{FAIRnets} \textit{FAIR}. This was achieved by following the \textit{FAIRification process}.\footnote{\url{https://www.go-fair.org/fair-principles/fairification-process/}, last acc. 2020-10-15.}
Our FAIRification process is aligned with the \textit{FAIRMetrics}\footnote{\url{https://fairmetrics.org}, last acc. 2020-10-15} outlined in Tab.~\ref{tab:fairmetrics}. In the following, we point out how the single FAIR metrics are met by our knowledge graph.

\noindent\textbf{Findable}
describes the property that metadata for digital assets is easy for both humans and machines to find.
Our approach ensured that, firstly, by retrieving the metadata available in the repository, secondly, structuring its metadata in the readme file, and thirdly, obtaining the architecture information from the code file according to the \textsc{FAIRnets Ontology}. The neural networks we model have unique identifiers (i.e., fulfilling \textit{Gen2\_FM\_F1A}) and a persistent URI (\textit{Gen2\_FM\_F1B}).
As a result, the process for a human to find a suitable neural network through resource identifiers in the metadata (\textit{Gen2\_FM\_F3}) is improved.
By using RDF as the data model and by providing a schema in OWL as well as a VoID file as a description of the knowledge graph, the metadata is machine-readable (\textit{Gen2\_FM\_F2}). Thus, the knowledge graph can be automatically filtered and used by services. An exemplary service supporting this statement is presented in Sec.~\ref{FAIRnetsSearch}.
\textsc{FAIRnets} allows for querying information about and within the architecture of the neural networks which was not possible previously. Now, complex queries are feasible (e.g., list all recurrent neural networks published in 2018), which cannot be solved by traditional keyword searches.
The metric \textit{Gen2\_FM\_F4}\footnote{\url{https://github.com/FAIRMetrics/Metrics/blob/master/FM_F4}, last acc. 2020-10-15} -- `indexed in a searchable resource' -- was not passed by \textsc{FAIRnets} although we indexed it on Zenodo. The reason is that the resource on Zenodo is not findable in the search engine Bing which the authors of the FAIRMetrics use as ground truth. However, \textsc{FAIRnets} is indexed by the search engine Google.

\noindent\textbf{Accessible}
describes that users can access (meta)data using a standardized communication protocol. The protocol must be open, free, and universally implemented. \textsc{FAIRnets Ontology and knowledge graph} is located on a web server and can be accessed using the HTTPS protocol (\textit{Gen2\_FM\_A1.1}). The neural networks in the repositories can also be accessed using the HTTPS protocol (\textit{Gen2\_FM\_A1.2}).
In addition to the open protocol, the accessible property requires that metadata can be retrieved, even if the actual digital assets are no longer available. Due to the separation of the information in \textsc{FAIRnets} and the actual neural networks on GitHub, this property is fulfilled, since the information in \textsc{FAIRnets} is preserved even if the neural networks on GitHub are no longer available (\textit{Gen2\_FM\_A2}).
The service to evaluate the metric \textit{Gen2\_FM\_A2} -- `metadata longevity' -- could not be executed because it only tests files that are less than 300kb\footnote{\url{https://github.com/FAIRMetrics/Metrics/blob/master/FM_A2}, last acc. 2020-10-15} whereas \textsc{FAIRnets} has more than 80MB. This test checks for the existence of the `persistence policy' predicate. This predicate is available in \textsc{FAIRnets}, which should pass the test.

\noindent\textbf{Interoperable}
refers to the capability of being integrated with other data as well as being available to applications for analysis, storage, and further processing.
We make use of Linked Data by applying RDF (\textit{Gen2\_FM\_I1}) and SPARQL to represent the information. This makes the data machine-readable, even without the specification of an ad-hoc algorithm or mapping. Additionally, the \textsc{FAIRnets Ontology} and the respective \textsc{knowledge graph} use well-established and commonly used vocabularies to represent the information. Among others, Dublin Core, Vocabulary of a Friend (VOAF), Creative Commons (CC), and a vocabulary for annotating vocabulary descriptions (VANN) are used for annotations and descriptions (\textit{Gen2\_FM\_I2}). 
As a further requirement of the FAIR guideline, qualified references to further metadata are required. This requirement is fulfilled by \texttt{rdfs:seeAlso} and \texttt{dc:references} (\textit{Gen2\_FM\_I3}). \texttt{dc:references} statements provide scientific references between the neural networks and scientific contributions. These references to the scientific contributions are provided via globally unique and persistent identifiers, such as DOIs.

\noindent\textbf{Reusable}
aims at achieving well-defined digital assets. This facilitates the replicability and usage in other contexts (i.e., reproducibility), as well as findability.
Due to the architecture and metadata extraction, the process of finding and reusing a neural network by an end-user becomes significantly easier and can now be performed systematically. By using best practices in ontology building, the properties and classes of \textsc{FAIRnets Ontology} provided are self-explanatory with labels and descriptions (\textit{Gen2\_FM\_R1.3}). The neural networks in \textsc{FAIRnets} contain structured detailed metadata such as creator and GitHub link (see \textit{Gen2\_FM\_R1.2}) for easy findability and reuse. At the same time, most neural networks in \textsc{FAIRnets} have an assigned license which is important for reusability (\textit{Gen2\_FM\_R1.1}).
For passing \textit{Gen2\_FM\_R1.2}, (meta)data must be associated with detailed provenance reusing existing vocabularies such as Dublin Core which we included in our knowledge graph. \textit{Gen2\_FM\_R1.3} tests a certification saying that the resource is compliant with a minimum of metadata. \textsc{FAIRnets} is described by using LOV standards for publication. Therefore, we assume that these metrics are fulfilled.
Overall, the neural networks modeled in \textsc{FAIRnets} fulfill all requirements of the FAIR principles, see Tab.~\ref{tab:fairmetrics}. 

\section{Impact}\label{sec:impact}

We see high potential of \textsc{FAIRnets Ontology} and the knowledge graph \textsc{FAIRnets} in the areas of \textit{transparency}, \textit{recommendation}, \textit{reusability}, \textit{education}, and \textit{search}. 
In the following, we outline these application areas in more detail.

\paragraph{Transparency.} Neural networks are applied in many different areas such as finance \cite{QiJLQ20}, medical health \cite{khan2001}, and law \cite{PALOCSAY2000271}. Transparency plays a major role in these areas when it comes to trust the output of a used neural network model. We claim that our contribution which makes neural networks more transparent can increase trust and privacy \cite{Schwabe19}. Additionally, using semantic annotations can even enhance interpretability by distributional semantics \cite{Silva2019}.

Another aspect is the transparency of scientific work regarding neural networks. Researchers publishing a model should provide it according to the FAIR principles to strengthen their scientific contribution. Our knowledge graph \textsc{FAIRnets} can pave the way for this.

\paragraph{Recommendation.} 
Neural Architecture Search (NAS) is used to find the best suitable neural network architecture based on existing architectures \cite{ElskenMH19}. However, the search is performed purely based on metrics like accuracy ignoring explainability aspects concerning the best fitting model. Our knowledge graph allows us to have a search for the best suitable neural network models on a meta-level, using modeled use-cases, data sets, and scientific papers. 
Knowledge graphs have also been used to provide explanations for recommendation to the user \cite{Wang00LC19,XianFMMZ19}. 

Additionally, we can apply explainable reasoning \cite{WangWX00C19} given the ontology and the knowledge graph and infer some rules. Doing this, we might reason which neural network models are reasonable or which components of the architecture stand in conflict with each other.

\paragraph{Reusability.} 
Transfer learning is a method in deep learning to reuse pre-trained models on new tasks. Our contribution facilitate the search of pre-trained neural networks and provide metadata needed to choose a specific neural network.
We can envision \textsc{FAIRnets} linked with other knowledge bases to enrich reusability of neural networks by applying Linked Data Principles.\textsuperscript{\ref{linkeddata}} For example, training data sets can be linked with Neural Data Server,\footnote{\url{http://aidemos.cs.toronto.edu/nds/}, last acc. 2020-10-15} Wikidata,\footnote{\url{https://www.wikidata.org}, last acc. 2020-10-15} and Zenodo\footnote{\url{https://zenodo.org}, last acc. 2020-10-15} through schema.org,\footnote{\url{https://schema.org}, last acc. 2020-10-15} scientific papers can be linked with the Microsoft Academic Knowledge Graph \cite{Farber19}, and metadata can be extended with OpenAIRE.\footnote{\url{https://www.openaire.eu}, last acc. 2020-10-15}

On the other hand, providing a model and encouraging its reuse can improve it by revealing limitations, errors, or suggestions to other tasks.

\paragraph{Education.}
Our \textsc{FAIRnets} knowledge graph can be used for educational purposes \cite{ChenLZCY18}, for instance, to learn best practices regarding designing a neural network model. Another aspect is to learn the usages of different architectures and their approaches (e.g., via linked papers). Our knowledge graph includes training parameters that can help setting up the training process of a neural network (e.g., when facing the cold start problem).

\label{FAIRnetsSearch}
\paragraph{Search}
We provide online the search system \textsc{FAIRnets Search}\footnote{\url{https://km.aifb.kit.edu/services/fairnets/}, last acc. 2020-10-15}~\cite{NguyenW19}, which is based on the proposed \textsc{FAIRnets Ontology} and \textsc{knowledge graph}.
Users can search for neural network models through search terms.
Additional information can be retrieved by using SPARQL as query language on top of our knowledge graph, which enables faceted and semantic search capabilities. The SPARQL endpoint is also available to the public. 
The search system shows how a semantic search system can be realized which improved the limited capabilities of keyword searches on GitHub.
Furthermore, developers can provide their GitHub repository to run the FAIRification process on their neural networks.
Until now, we have over 550 visits to the website \textsc{FAIRnets Search} with over 4,800 page views, 1,400 searches on our website with an average duration of twelve minutes, and the maximal actions in one visit is 356.

\section{Related Work}
\label{sec:related_work}

\paragraph{Information of neural network models.}
Mitchell et al.~\cite{mitchell_model_2019} suggest which information about neural networks should be considered as relevant when modeling them. Information such as description, date of the last modification, link to papers, or other resources to further information, as well as the intended purpose of a neural network, are taken into account.
Storing such information makes the neural networks more transparent.
We follow this suggestion by defining a semantic representation which, to the best of our knowledge, does not exist for neural network models so far.

The knowledge extraction from neural networks can point out relevant features or redundancies \cite{BG1997}. We extract neural network information to build a knowledge graph to better evaluate the causal relationships between different neural network architectures.

\paragraph{Representing and provisioning neural network models.}
There exist several standards for the exchange of neural network information on the instance level.
The Predictive Model Markup Language (PMML)\footnote{\url{http://dmg.org/pmml/v4-0-1/NeuralNetwork.html}, last acc. 2020-10-15} is an XML-based standard for analytic models developed by the Data Mining Group. 
PMML is currently supported by more than 30 organizations.
The Open Neural Network eXchange format (ONNX)\footnote{\url{https://onnx.ai}, last acc. 2020-10-15}
is a project by Facebook and Microsoft that converts neural networks into different frameworks. 
These two formats serve as an exchange format for neural networks on instance level. 
We are less interested in the exchange of formats, but rather the reusability of the neural networks on a meta-level.
Therefore, our \textsc{FAIRnets Ontology} lifts it's elements to a semantic level, i.e. to RDF/S, following a methodology for reusing ontologies~\cite{pinto_reusing_nodate} and applying the Linked Data Principles.\textsuperscript{\ref{linkeddata}}
Thus, we incorporate information on the instance and meta-level in the knowledge graph \textsc{FAIRnets}.

\paragraph{Neural network repositories.}
Many pre-trained neural networks are available online.
The well-known Keras framework offers ten pre-trained neural networks for reuse.\textsuperscript{\ref{keras}}
The Berkeley Artificial Intelligence Research Lab has a deep learning framework called Caffe Model Zoo\textsuperscript{\ref{modelzoo}} which consists of about fifty neural networks.
Wolfram Alpha has a repository with neural networks\textsuperscript{\ref{wolfram}} which consists of approximately ninety models. These pre-trained neural networks are represented in different formats making it, for instance, difficult to compare or reuse neural networks. 
Besides, a larger number of neural networks can be found in code repositories such as GitHub. 
These neural networks are typically coded in one of the major programming frameworks such as Keras, TensorFlow, or PyTorch.
Our approach aims to consider such neural networks and make them available as FAIR data. 
\section{Conclusion and Future Work}
\label{sec:conclusion}

This paper was dedicated to make neural networks FAIR. To this end, we first proposed the \textsc{FAIRnets Ontology}, an ontology that allows us to model neural networks on a fine-grained level and that is easily extensible. Second, we provided the knowledge graph \textsc{FAIRnets}. This graph contains rich metadata of 18,463 publicly available neural network models using our proposed ontology as knowledge schema. Both the \textsc{FAIRnets Ontology} as well as the \textsc{FAIRnets} knowledge graph show a high potential impact in fields like recommender systems and education. 

For the future, we plan to connect the \textsc{FAIRnets Ontology} and \textsc{knowledge graph} with scholarly data. Specifically, we will work on linking publications, authors, and venues modeled in knowledge graphs like the Microsoft Academic Knowledge Graph~\cite{Farber19} or Wikidata to the \textsc{FAIRnets} knowledge graph. This will require to apply sophisticated information extraction methods on scientific publications.

\bibliographystyle{splncs04}
\bibliography{references}
\end{document}